\documentclass[10pt,twocolumn,letterpaper]{article}

\usepackage{cvpr}
\usepackage{times}
\usepackage{epsfig}
\usepackage{graphicx}
\usepackage{amsmath}
\usepackage{amssymb}
\usepackage{graphicx}

\usepackage{subcaption}
\usepackage[linesnumbered,ruled]{algorithm2e}
\usepackage{setspace}
\usepackage{xspace}
\usepackage{enumitem}
\usepackage{multirow}
\usepackage{pifont}
\usepackage{titling}
\usepackage{stfloats}
\newcommand{\R}{$R_{\epsilon}$\xspace}
\newcommand{\DP}{$P_d$\xspace}
\newcommand{\DR}{$r_d$\xspace}
\newcommand{\sads}{SADS\xspace}

\usepackage[pagebackref=true,breaklinks=true,letterpaper=true,colorlinks,bookmarks=false]{hyperref}

\cvprfinalcopy 

\def\cvprPaperID{xxxx} 

\ifcvprfinal\fi
\begin{document}

\title{Single-step  Adversarial  training  with Dropout Scheduling}

\author{Vivek B.S. and R. Venkatesh Babu\\
Video Analytics Lab, Department of Computational and Data Sciences\\
Indian Institute of Science, Bangalore, India 
}
\predate{}
\postdate{}
\date{}
\maketitle
\thispagestyle{empty}

\begin{abstract}
Deep learning models have shown impressive performance across a spectrum of computer vision applications including medical diagnosis and autonomous driving. One of the major concerns that these models face is their susceptibility to adversarial attacks. Realizing the importance of this issue, more researchers are working towards developing robust models that are less affected by adversarial attacks. Adversarial training method shows promising results in this direction. In adversarial training regime, models are trained with mini-batches augmented with adversarial samples. Fast and simple methods (e.g., single-step gradient ascent) are used for generating adversarial samples, in order to reduce computational complexity. It is shown that models trained using single-step adversarial training method (adversarial samples are generated using non-iterative method) are pseudo robust. Further, this pseudo robustness of models is attributed to the gradient masking effect. However, existing works fail to explain \textit{when and why gradient masking effect occurs during single-step adversarial training}. In this work, (i) we show that models trained using single-step adversarial training method learn to prevent the generation of single-step adversaries, and this is due to over-fitting of the model during the initial stages of training, and (ii) to mitigate this effect, we propose a single-step adversarial training method with dropout scheduling. Unlike models trained using existing single-step adversarial training methods, models trained using the proposed single-step adversarial training method are robust against both single-step and multi-step adversarial attacks, and the performance is on par with models trained using computationally expensive multi-step adversarial training methods, in white-box and black-box settings.
\end{abstract}
\section{Introduction}
\label{sec:Introduction}
Machine learning models are susceptible to adversarial samples: samples with imperceptible, engineered noise designed to manipulate model's output~\cite{adversarialml-acmmm-2011,evasion-mlkd-2013,intriguing-iclr-2014,prsystemsunderattack-pari-2014,explainingharnessing-iclr-2015,limitations-eurosp-2016}. Further, Szegedy~\etal\cite{intriguing-iclr-2014} observed that these adversarial samples are transferable across multiple models i.e., adversarial samples generated on one model might mislead other models. Due to which, models deployed in the real world are susceptible to  black-box attacks~\cite{delving-iclr-2017,practicalbb-asiaccs-2017}, where limited or no knowledge of the deployed model is available to the attacker. Various schemes have been proposed to defend against adversarial attacks (e.g.,~\cite{explainingharnessing-iclr-2015,defensivedistillation-arxiv-2015,ondetecting-iclr-2017}), in this direction \textit{Adversarial Training (AT)} procedure~\cite{explainingharnessing-iclr-2015,ensembleAT-iclr-2018,madry-iclr-2018,trades_icml} shows promising results. 

In  adversarial training regime, models are trained with mini-batches containing adversarial samples typically generated by the model being trained. Adversarial sample generation methods range from simple methods~\cite{explainingharnessing-iclr-2015} to complex optimization methods~\cite{deepfool-cvpr-2016}. In order to reduce computational complexity, non-iterative methods such as Fast Gradient Sign Method (FGSM)~\cite{explainingharnessing-iclr-2015}  are typically used for generating adversarial samples. Further, it has been shown that models trained using single-step adversarial training methods are pseudo robust~\cite{ensembleAT-iclr-2018}:
\begin{itemize}
    \item Although these models appears to be robust to single-step attacks in white-box setting (complete knowledge of the deployed model is available to the attacker), they are susceptible to single-step attacks (non-iterative methods) in black-box attack setting~\cite{ensembleAT-iclr-2018}.
    \item Further, these models are susceptible to multi-step attacks (iterative methods) in both white-box setting~\cite{atscale-iclr-2017} and black-box setting~\cite{dong2018boosting}.
\end{itemize}
Tramer~\etal\cite{ensembleAT-iclr-2018} demonstrated that  models trained using single-step adversarial training method converges to degenerative minima, and exhibit \textit{gradient masking effect}. Single-step adversarial sample generation methods such as FGSM, compute adversarial perturbations based on the linear approximation of the model's loss function i.e., image is perturbed in the direction of the gradient of loss with respect to the input image. Gradient masking effect causes this linear approximation of loss function to become unreliable for generating adversarial samples during single-step adversarial training. Madry~\etal\cite{madry-iclr-2018} demonstrated that models trained using adversarial samples that maximize the training loss are robust against single-step and multi-step attacks.  Such samples could be generated using the Projected Gradient Descent (PGD). However, PGD method is an iterative method, due to which training time increases substantially. Though prior works have enabled to learn robust models, they fail to answer the following important questions: (i)  \textit{Why models trained using single-step adversarial training method exhibit gradient masking effect?} and (ii) \textit{At what phase of the single-step adversarial training, the model starts to exhibit gradient masking effect?}

In this work, we attempt to answer these questions and propose a novel single-step adversarial training method to learn robust models. First, we show that models trained using single-step adversarial training method learn to prevent the generation of single-step adversaries, and this is due to over-fitting of the model during the initial stages of training. Over-fitting of the model on single-step adversaries causes linear approximation of loss function to become unreliable for generating adversarial samples i.e., gradient masking effect. Finally, we propose a single-step adversarial training method with dropout scheduling to learn robust models. Note that, just adding dropout layer (typical setting: dropout layer with fixed dropout probability after FC+ReLU layer) does not help the model trained using single-step adversarial training method to gain robustness. Prior works observed no significant improvement in the robustness of models (with dropout layers in typical setting),  trained using normal training and single-step adversarial training methods~\cite{explainingharnessing-iclr-2015,atscale-iclr-2017}. Results for these settings are shown in section~\ref{subsec:dropout_effect}. Unlike typical setting, we introduce dropout layer after each non-linear layer (i.e., dropout-2D after conv2D+ReLU, and dropout-1D after FC+ReLU) of the model, and further decay its dropout probability as training progress. Interestingly, we show that this proposed dropout setting has significant impact on the model's robustness. 
The major contributions of this work can be listed as follows: 
\begin{itemize}
    \item We show that models trained using single-step adversarial training method learns to prevent the generation of single-step adversaries, and this is due to over-fitting of the model during the initial stages of training.
    \item  Harnessing on the above observation, we propose a single-step adversarial training method with dropout probability scheduling. Unlike models trained using existing single-step adversarial training methods, models trained using the proposed method are robust against both single-step and multi-step attacks. 
    \item The proposed single-step adversarial training method is much faster than multi-step adversarial training methods, and achieves on par results.  
\end{itemize}
\vspace{-0.3cm}
\section{Notations}
\label{sec:notations} 
Consider a neural network $f$ trained to perform image classification task, and $\theta$ represents parameters of the neural network. Let $x$ represents the image from the dataset and $y_{true}$ be its corresponding ground truth label. The neural network is trained using loss function $J$ (e.g., cross-entropy loss), and $\nabla_{x}J$ represents the gradient of loss with respect to the input image $x$. Adversarial image $x_{adv}$ is generated by adding norm-bounded perturbation $\delta$  to the image $x$. Perturbation size ($\epsilon$) represents the $l_{\infty}$ norm constraint on the generated adversarial perturbation i.e., $||\delta||_{\infty}\le\epsilon$. Please refer to supplementary document for details on adversarial training and  attack generation methods.
\section{Related Works}
\label{sec:Related Works} 
Following the findings of Szegedy~\etal\cite{intriguing-iclr-2014}, various  attacks (e.g.,~\cite{explainingharnessing-iclr-2015,deepfool-cvpr-2016,robustness-arxiv-2016,mopuri-bmvc-2017,mopuri2019generalizable,dong2018boosting,ganeshan2019fda} have been proposed. Further, in order to defend against adversarial attacks, various schemes such as adversarial training (e.g.,~\cite{explainingharnessing-iclr-2015,atscale-iclr-2017,madry-iclr-2018,trades_icml,vivek2018gray,S_2019_CVPR_Workshops}) and input pre-processing (e.g.,~\cite{guo2018countering,samangouei2018defensegan}) have been proposed. Athalye~\etal\cite{obfuscated-gradients} showed that obfuscated gradients give a false sense of robustness, and broke seven out of nine defense papers~\cite{buckman2018thermometer,ma2018characterizing,guo2018countering,xie2018mitigating,song2018pixeldefend,samangouei2018defensegan,madry-iclr-2018,ma2018characterizing,s.2018stochastic} accepted to ICLR 2018.~In this direction, adversarial training method~\cite{madry-iclr-2018}, shows promising results for learning
robust deep learning models. Kurakin~\etal\cite{atscale-iclr-2017} observed that models trained using single-step adversarial training methods are susceptible to multi-step attacks. Further, Tramer~\etal\cite{ensembleAT-iclr-2018} demonstrated that these models exhibit gradient masking effect, and proposed Ensemble Adversarial Training (EAT) method. However, models trained using EAT are still susceptible to multi-step attacks in white-box setting. Madry~\etal\cite{madry-iclr-2018} demonstrated that adversarially trained model can be made robust against white-box attacks, if  perturbation crafted while training  maximizes the loss. Zhang~\etal\cite{trades_icml} proposed a regularizer for multi-step adversarial training, that encourages the output of the network to be smooth. On the other hand, works such as~\cite{raghunathan2018certified} and~\cite{wong2017provable} propose a method to learn models that are provably robust against norm bounded adversarial attacks. However, scaling these methods to deep networks and  large perturbation sizes is difficult. Whereas, in this work we show that it is possible to learn robust models using single-step adversarial training method, if over-fitting of the model on adversarial samples is prevented during training. We achieve this by introducing dropout layer after each non-linear layer of the model with a dropout schedule.
\begin{figure*}[h!]
    \centering
	\includegraphics[width=0.99\linewidth]{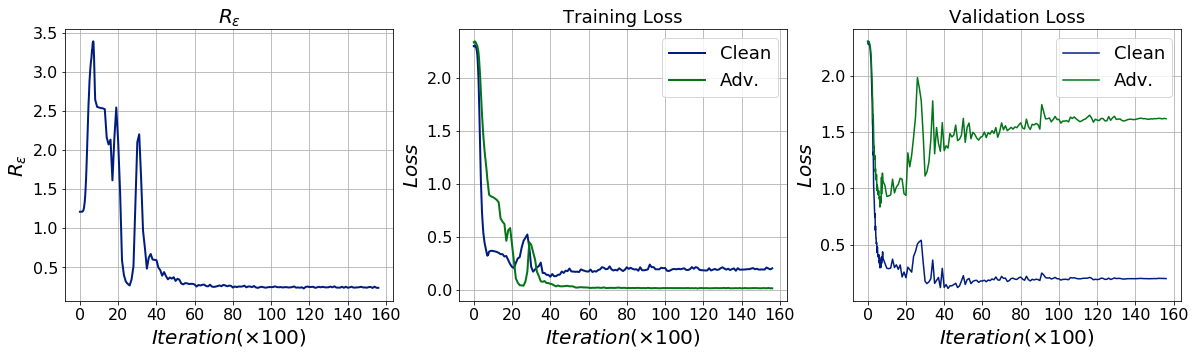}
    \vspace{-0.3cm}
	\caption{\textbf{Single-step adversarial training:} Trend of $R_{\epsilon}$, training loss, and validation loss during single-step adversarial training, obtained for LeNet+ trained on MNIST dataset. Column-1: plot of $R_{\epsilon}$ versus training iteration. Column-2: training loss versus training iteration. Column-3: validation loss versus training iteration. Note that, when $R_{\epsilon}$ starts to decay, loss on adversarial validation set starts to increase  indicating that the model is over-fitting on the adversarial samples.}
	\label{fig:trends_during_fgsm_training}
\end{figure*}
\begin{figure*}[h!]
\vspace{-0.2cm}
    \centering
	\includegraphics[width=0.9\linewidth]{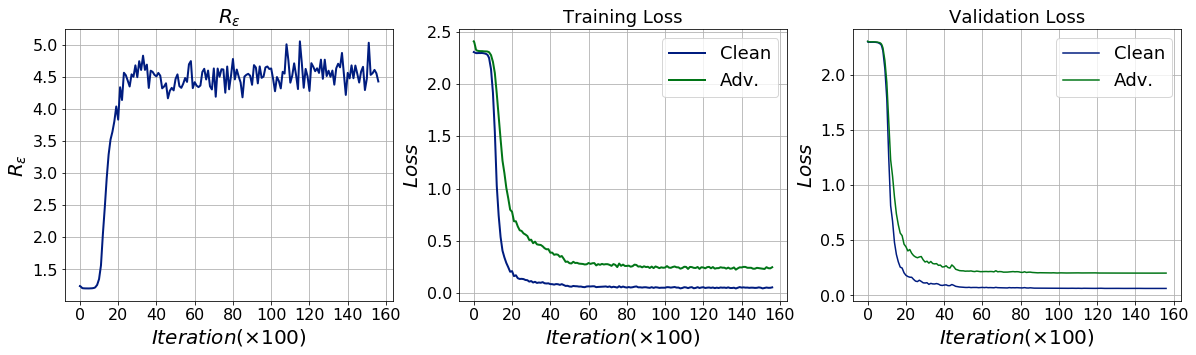}
    \vspace{-0.3cm}
	\caption{\textbf{Multi-step adversarial training:} Trend of $R_{\epsilon}$, training loss, and validation loss during multi-step adversarial training, obtained for LeNet+ trained on MNIST dataset. Column-1: plot of $R_{\epsilon}$ versus training iteration. Column-2: training loss versus training iteration. Column-3: validation loss versus training iteration. Note that, for the entire training duration $R_{\epsilon}$ does not decay, and no over-fitting effect can be observed.} 
	\label{fig:trends_during_pgd_training}  
\vspace{-0.4cm}
\end{figure*}
\section{Over-fitting and its effect during adversarial training}
\label{sec:overfitting_adv_training}
In this section, we show that models trained using single-step adversarial training method learn to prevent the generation of single-step adversaries, and this is due to over-fitting of the model during the initial stages of training. First, we discuss the criteria for learning robust models using adversarial training method, and then we show that this criteria is not satisfied during single-step adversarial training method. Most importantly, we show that over-fitting effect is the reason for failure to satisfy the criteria.

Madry~\etal\cite{madry-iclr-2018} demonstrated that it is possible to learn robust models using adversarial training method, if adversarial perturbations ($l_{\infty}$ norm bounded) crafted while training maximizes the model's loss.  This training objective is formulated as a minimax optimization problem (Eq.~\ref{equ:pgd_formaulation}). Where $\psi$ represents the feasible set e.g., for $l_\infty$ norm constraint attacks $\psi=\{\delta:||\delta||_{\infty}\le\epsilon\}$, and $D$ is the training set.
\begin{eqnarray}
\label{equ:pgd_formaulation}
&\displaystyle{\min_{\theta} \bigg[E_{(x,y)\in D}\big[\displaystyle{\max_{\delta\in \psi}}~J\big(f(x+\delta;\theta),y_{true}\big)\big]\bigg]}
\end{eqnarray}
\begin{equation} 
\label{equ:loss_ratio}
R_{\epsilon}= \frac{loss_{adv}}{loss_{clean}}
\end{equation}
 At each iteration, norm bounded adversarial perturbations  that maximizes the training loss should be generated. Further, the model's parameters ($\theta$) should be updated so as to decrease the loss on such adversarial samples. Madry~\etal\cite{madry-iclr-2018} solves the maximization step by generating adversarial samples using an iterative method named Projected Gradient Descent (PGD). In order to quantify the extent of inner maximization of Eq.~(\ref{equ:pgd_formaulation}), we compute loss ratio \R using Eq.~(\ref{equ:loss_ratio}).  Loss ratio  is defined as the ratio of loss on the adversarial samples to the loss on its corresponding clean samples for a given perturbation size $\epsilon$. The metric \R captures the extent of inner maximization achieved by the generated adversarial samples i.e., factor by which loss has increased by perturbing the clean samples.\\
A sample is said to be an adversarial sample if it is capable of manipulating the model's prediction. Such manipulations could be achieved by perturbing the samples along the adversarial direction~\cite{explainingharnessing-iclr-2015}. A perturbation is said to be an \textit{adversarial perturbation} when it causes loss on the perturbed sample to increase. This implies that the loss on the adversarially perturbed samples should be greater than the loss on the corresponding unperturbed samples i.e., $loss_{adv}>loss_{clean}$. Based on these facts, \R can be interpreted in the following manner:
\begin{itemize}
    \item Generated perturbation is said to be an \textit{adversarial perturbation} if  \R$>$1 i.e., $loss_{adv}>loss_{clean}$  
    \item \R$<$1 i.e., $loss_{adv}<loss_{clean}$, implies that the generated perturbation is not an adversarial perturbation. The attack method fails to generate \textit{adversarial perturbations} for the given model. 
\end{itemize}
We obtain the plot of \R versus iteration for models trained using single-step adversarial training method~\cite{explainingharnessing-iclr-2015} and multi-step adversarial training method~\cite{madry-iclr-2018}. Column-1 of Fig.~\ref{fig:trends_during_fgsm_training} and Fig.~\ref{fig:trends_during_pgd_training} show these plots obtained for LeNet+ trained on MNIST dataset~\cite{lecun1998mnist} using single-step and multi-step adversarial training methods respectively. It can be observed that during single-step adversarial training, \R initially increases and then starts to decay rapidly. Further \R becomes less than one after 20 ($\times$100) iterations. This implies that single-step adversarial sample generation method is unable to generate \textit{adversarial perturbations} for the model, leading to adversarial training without useful adversarial samples.

We demonstrate this behavior of the model to prevent the inclusion of adversarial samples is due to over-fitting on the adversarial samples. Typically during normal training, loss on the validation set is monitored to detect over-fitting effect i.e., validation loss increases when the model starts to over-fit on the training set. Unlike normal training, during adversarial training we monitor the loss on the clean and adversarial validation set. A normally trained model is used for generating adversarial validation set, so as to ensure that the generated adversarial validation samples are independent of the model being trained. Column-2 and column-3 of Fig.~\ref{fig:trends_during_fgsm_training} shows the plot of loss versus iteration during training of LeNet+ on MNIST dataset using single-step adversarial training. It can be observed that, when \R starts to decay, loss on the adversarial validation set starts to increase. This increase in the validation loss indicates over-fitting of the model on the single-step adversaries. Whereas, during multi-step adversarial training method, \R initially increases and then saturates (column-1, Fig.~\ref{fig:trends_during_pgd_training}). Further, no such over-fitting effect is observed for the entire training duration (column-3, Fig.~\ref{fig:trends_during_pgd_training}). Note that, a normally trained model was used for generating FGSM ($\epsilon$=0.3) adversarial validation set, and we observe similar trend if a normally trained model of different architecture is used for generating FGSM adversarial validation set, please refer to supplementary document.
\subsection{Effect of dropout layer}
\label{subsec:dropout_effect}
\begin{figure}[t!]
    \centering
    \includegraphics[width=0.99\linewidth]{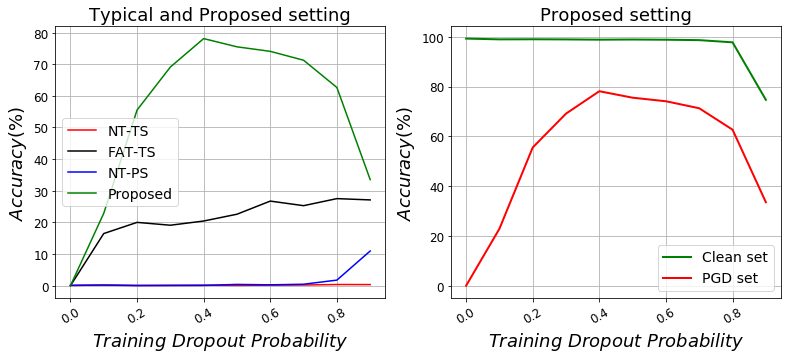}
    \vspace{-0.4cm}
    \caption{\textbf{Column-1:} Effect of dropout probability of dropout layers in typical setting and in the proposed setting on the model's robustness against PGD attack ($\epsilon$=0.3, $\epsilon_{step}$=0.01 and $steps$=40). Obtained for LeNet+ trained on MNIST dataset. NT-TS: Normal training with dropout layer in typical setting. FAT-TS: Single-step adversarial training with dropout layer in typical setting. NT-PS: Normal training with dropout layer in the proposed setting. Proposed: Single-step adversarial training with dropout layer in the proposed setting. \textbf{Column-2:} Effect of dropout probability  on the model's accuracy on clean and PGD adversarial validation set ($\epsilon$=0.3, $\epsilon_{step}$=0.01 and $steps$=40). Obtained for LeNet+ with dropout layer in the proposed setting, trained using single-step adversarial training method on MNIST dataset.}
    \label{fig:Dropout_Effect}  
\end{figure}

In the previous section, we showed that models trained using single-step adversarial training learn to prevent the generation of single-step adversaries. Further, we demonstrated that this behavior of models is due to over-fitting.
Dropout layer~\cite{srivastava2014dropout} has been shown to be effective in mitigating over-fitting during training, and typically dropout-1D layer is added after FC+ReLU layers in the networks. We refer to this setting as \textit{typical setting}. Prior works which used dropout layer during single-step adversarial training observed no significant improvement in the model's robustness. This is due to the use of dropout layer in \textit{typical setting}. Whereas, we empirically show that it is necessary to introduce dropout layer after every non-linear layer of the model (\textit{proposed dropout setting} i.e., dropout-2D after Conv2D+ReLU layer and dropout-1D after FC+ReLU layer) to  mitigate over-fitting during single-step adversarial training, and to enable the model to gain robustness against adversarial attacks (single-step and multi-step attacks). We train LeNet+ with dropout layer in \textit{typical setting} and in the \textit{proposed setting} respectively, on MNIST dataset using single-step adversarial training method for different values of dropout probability. After training, we obtain the performance of these resultant models against PGD attack ($\epsilon$=0.3, $\epsilon_{step}$=0.01, $steps$=40). Column-1 of Fig.~\ref{fig:Dropout_Effect} shows the trend of accuracy of these models for PGD attack with respect to the dropout probability used while training. It can be observed that the gain in the robustness of adversarially trained model with dropout layer in the proposed setting is significantly better compared to the adversarially trained model with dropout layer in typical setting (FAT-TS). From column-2 of Fig.~\ref{fig:Dropout_Effect}, it can be observed that the robustness of adversarially trained model with dropout layer in the proposed setting, increases with the increase in the dropout probability ($p$) and reaches a peak value at $p$=0.4. Further increase in the dropout probability causes decrease in the accuracy  on both clean and adversarial samples.  Based on this observation, we propose an improved single-step adversarial training in the next subsection. Furthermore, we perform normal training of LeNet+ with dropout layers in typical setting and in the proposed setting, on MNIST dataset. From column-1 of Fig.~\ref{fig:Dropout_Effect}, it can be observed that there is no significant improvement in the robustness of these normally trained models.
\subsection{SADS: Single-step Adversarial training with Dropout Scheduling}
\label{subsec:DropAdv}
 Column-1 of Fig.~\ref{fig:Dropout_Effect} indicates that use of dropout layer in typical setting is not sufficient to avoid over-fitting on adversarial samples, and we need severe dropout regime involving all the layers (i.e., proposed setting: dropout layer after Conv2D+ReLU and FC+ReLU layers) of the network in order to avoid over-fitting. For the proposed dropout regime, determining exact dropout probability is network dependent and is difficult. Further, having high dropout probability causes under-fitting of the model, and having low dropout probability causes the model to over-fit on the adversarial samples. 

Based on these observations, we propose a single-step adversarial training method with dropout scheduling (Algorithm~\ref{alg:ProposedAlgorithm}). In the proposed training method, we introduce dropout layer after each non-linear layer  of the model to be trained. We initialize these dropout layers with a high dropout probability \DP. Further, during training we linearly decay the dropout probability of all the dropout layers and this decay in the dropout probability is controlled by the hyper-parameter \DR. The hyper-parameter, \DR is expressed in terms of maximum training iterations (e.g., \DR=$1/2$ implies that dropout probability reaches zero when the current training iteration is equal to half of the maximum training iterations). In experimental section~\ref{sec:Experiments}, we show the effectiveness of the proposed training method. Note that dropout layer is only used while training. 
\section{Experiments}
\label{sec:Experiments}
In this section, we show the effectiveness of models trained using the proposed single-step adversarial training method (SADS) in white-box and black-box settings. We perform the sanity tests described in~\cite{carlini2019evaluating}, in order to verify that models trained using \sads are robust and does not exhibit \textit{obfuscated gradients} (Athalye~\etal\cite{obfuscated-gradients} demonstrated that models exhibiting obfuscated gradients are not robust against adversarial attacks). We show results on MNIST~\cite{lecun1998mnist}, Fashion-MNIST~\cite{fashion_mnist} and CIFAR-10~\cite{cifar_10_dataset} datasets. We use LeNet+ (please refer to supplementary document for details on network architecture) for both MNIST and Fashion-MNIST datasets. For CIFAR-10 dataset, WideResNet-28-10~\cite{BMVC2016_87} is used. These models are trained using SGD with momentum. Step-policy is used for learning rate scheduling. For all datasets, images are pre-processed to be in [0,1] range. For CIFAR-10, random crop and horizontal flip are performed for data-augmentation.
\begin{algorithm}[t!]
\setstretch{1.00}
\SetKwInOut{Input}{Input}
\Input{\newline
Training mini-batch size ($m$) \newline
Maximum training iterations ($Max_{itertion}$) \newline
Hyper-parameters: \DP, \DR
}
\textbf{Initialization} \newline
Randomly initialize network $N$\newline
$iteration = 0$\newline
$prob$ = \DP \newline
Insert dropout layer after each non-linear layer of the network $N$  \newline
Set dropout probability (p) of all the dropout layers with $prob$ \newline 
\While{$iteration \leq Max_{itertion}$}{
  Read minibatch $B$=$\{x^1,..,x^m\}$ from training set\\
  Compute FGSM adversarial sample $\{x_{adv}^1,...,x_{adv}^m\}$ from corresponding clean samples $\{x^1,...,x^m\}$ using the current state of the network $N$ \\
  Make new minibatch $B^{*}=\{x_{adv}^1,...,x_{adv}^m\}$ \newline
  \textcolor{blue}{/*Forward pass, compute loss, backward pass, and update parameters*/} \\
  Do one training step of Network $N$ using minibatch $B^{*}$ \newline
  \textcolor{blue}{/*Update dropout probability of Dropout-1D and Dropout-2D layers with $prob$*/}\\ 
  $prob$ = $max(\ 0,\ P_d\cdot( 1 - \frac{iteration}{r_d \cdot Max_{itertion}})\ )$\\
  $iteration = iteration + 1$ \\
 }
 \caption{Single-step Adversarial training with Dropout Scheduling (\sads)}
 \label{alg:ProposedAlgorithm}
\end{algorithm}
\begin{table}[t!]

    \centering
    \caption{\textbf{MNIST: White-Box setting.} Classification accuracy (\%) of models trained on MNIST dataset using different training methods. For all attacks $\epsilon$=0.3 is used and for PGD attack $\epsilon_{step}$=0.01 is used. For both IFGSM and PGD attacks, steps is set to 40.}
    \label{table:whitebox_mnist}
    \begin{tabular}{lrrrrr} \hline \hline 
    \multicolumn{2}{l}{\textbf{Training }} & \multicolumn{4}{c}{\textbf{Attack Method}} \\ \cline{4-6}
    \multicolumn{2}{l}{\textbf{Method}} & \textbf{Clean}  & \textbf{FGSM}  & \textbf{IFGSM}     & \textbf{PGD}    \\
     \hline \hline
    \multicolumn{2}{l}{NT}          & 99.24 & 11.65 & 0.31  & 0.01      \\ \hline 
    \multicolumn{6}{c}{Multi-step adversarial training}\\ \hline
    \multicolumn{2}{l}{PAT}         & 98.41 & 95.56 & 92.64 & 92.08     \\ 
    \multicolumn{2}{l}{TRADES}      & 98.70 & 96.30 & 95.14 & 95.05     \\ \hline
    \multicolumn{6}{c}{Single-step adversarial training}\\ \hline
    \multicolumn{2}{l}{FAT}         & 99.34 & 89.04 & 1.19  & 0.17     \\ 
    \multicolumn{2}{l}{\sads}       & 98.89 &	94.78 &	89.35 &	88.51     \\  
    \multicolumn{2}{l}{}            & $\pm$0.01  & $\pm$0.19 & $\pm$0.09 & $\pm$0.22     \\ \hline\hline
    \end{tabular}
\vspace{-0.1cm}
\end{table}
\begin{table}[t!]
    \centering
    \caption{\textbf{Fashion-MNIST: White-Box attack.} Classification accuracy (\%) of models trained on Fashion-MNIST dataset using different training methods. For all attacks $\epsilon$=0.1 is used and for PGD attack $\epsilon_{step}$=0.01 is used. For both IFGSM and PGD attacks, steps is set to 40.}
    \label{table:whitebox_fmnist}
    \begin{tabular}{lcrrrr} \hline \hline
    \multicolumn{2}{l}{\textbf{Training}} & \multicolumn{4}{c}{\textbf{Attack Method}} \\ \cline{4-6}
    \multicolumn{2}{l}{\textbf{Method}} & \textbf{Clean}  & \textbf{FGSM}  & \textbf{IFGSM}     & \textbf{PGD}         \\
     \hline \hline
    \multicolumn{2}{l}{NT}          & 91.42 & 6.46  & 1.01  & 0.16      \\ \hline
    \multicolumn{6}{c}{Multi-step adversarial training}\\ \hline    
    \multicolumn{2}{l}{PAT}         & 84.55 & 77.30 & 75.95 & 75.18   \\ 
    \multicolumn{2}{l}{TRADES}      & 86.69 & 80.39 & 78.94 & 78.04   \\ \hline 
    \multicolumn{6}{c}{Single-step adversarial training}\\ \hline   
    \multicolumn{2}{l}{FAT}         & 90.45 & 83.43 & 21.26 & 16.65     \\ 
    \multicolumn{2}{l}{\sads}       & 85.21 & 75.81 & 71.14 & 69.51    \\ 
    \multicolumn{2}{l}{}            & $\pm$0.08  & $\pm$1.31  & $\pm$1.01  & $\pm$1.43    \\ \hline\hline 
    \end{tabular}
\vspace{-0.1cm}
\end{table}
\\\textbf{Evaluation:} We show the performance of models against adversarial attacks  in white-box and black-box setting. For \sads, we report mean and standard deviation over three runs.
\\\textbf{Attacks:} For $l_{\infty}$ based attacks, we use Fast Gradient Sign Method (FGSM)~\cite{explainingharnessing-iclr-2015}, Iterative Fast Gradient Sign Method (IFGSM)~\cite{physicalworld-arxiv-2016}, Momentum Iterative Fast Gradient Sign Method (MI-FGSM)~\cite{dong2018boosting} and Projected Gradient Descent (PGD)~\cite{madry-iclr-2018}. For $l_2$ based attack, we use DeepFool~\cite{deepfool-cvpr-2016} and Carlini \& Wagner~\cite{robustness-arxiv-2016}. 
\\\textbf{Perturbation size:} For $l_{\infty}$ based attacks, we set perturbation size ($\epsilon$) to the values described in~\cite{madry-iclr-2018} i.e., $\epsilon$=0.3, 0.1 and 8/255 for MNIST, Fashion-MNIST and CIFAR-10 datasets respectively.
\\\textbf{Comparisons:} We compare the performance of the proposed single-step adversarial training method (\sads) with Normal training (NT), FGSM adversarial training (FAT)~\cite{atscale-iclr-2017}, Ensemble adversarial training (EAT)~\cite{ensembleAT-iclr-2018}, PGD adversarial training (PAT)~\cite{madry-iclr-2018}, and TRADES~\cite{trades_icml}. Note that, FAT, EAT and \sads (ours)  are single-step adversarial training methods, whereas PAT and TRADES are multi-step adversarial training methods.  Results for EAT are shown in supplementary document.
\begin{table}[h!]
    \centering
    \caption{\textbf{CIFAR-10: White-Box attack.} Classification accuracy (\%) of models trained on CIFAR-10 dataset using different training methods.  For all attacks $\epsilon$=8/255 is used and for PGD attack $\epsilon_{step}$=2/255 is used. For both IFGSM and PGD attacks, steps is set to 7.}
    \label{table:whitebox_cifar}
    \begin{tabular}{lcrrrr} \hline \hline
    \multicolumn{2}{l}{\textbf{Training}} & \multicolumn{4}{c}{\textbf{Attack Method}} \\ \cline{4-6}
    \multicolumn{2}{l}{\textbf{Method}} & \textbf{Clean}  & \textbf{FGSM}  & \textbf{IFGSM}     & \textbf{PGD}         \\
     \hline \hline
    \multicolumn{2}{l}{NT}          & 94.75 & 28.16 & 0.07  & 0.03  \\\hline
    \multicolumn{6}{c}{Multi-step adversarial training}\\ \hline 
    \multicolumn{2}{l}{PAT}         & 85.70 & 53.96 & 48.65 & 47.23 \\
    \multicolumn{2}{l}{TRADES}      & 87.20 & 56.34 & 51.21 & 50.03 \\\hline
    \multicolumn{6}{c}{Single-step adversarial training}\\ \hline 
    \multicolumn{2}{l}{FAT}         & 94.04 & 98.54 & 0.31  & 0.09  \\\hline
    \multicolumn{2}{l}{\sads}       & 82.01 & 51.99 & 46.37 & 45.66 \\ 
    \multicolumn{2}{l}{}            & $\pm$0.06  & $\pm$1.02  & $\pm$1.17  & $\pm$1.26    \\ \hline \hline
    \end{tabular}
\vspace{-0.1cm}
\end{table}
\subsection{Performance in White-box setting}
\label{subsec:exp_white_box_setting}
\begin{figure*}[t!]
\centering    
	\includegraphics[width=0.90\linewidth]{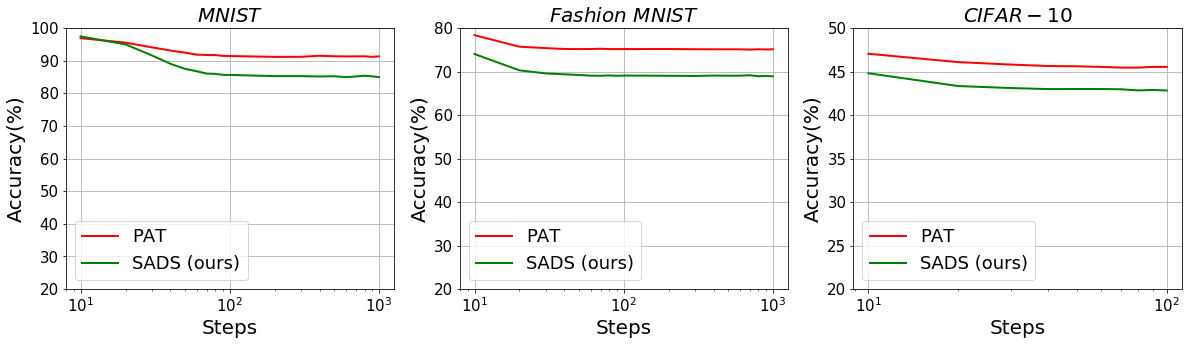}
	\vspace{-0.3cm}
	\caption{Plot of accuracy of the model trained using PAT and SADS, on PGD adversarial test set versus steps of PGD attack with fixed $\epsilon$. For PGD attack we set ($\epsilon$,$\epsilon_{step}$) to (0.3,0.01), (0.1,0.01) and (8/255,2/255) for MNIST, Fashion-MNIST and CIFAR-10 datasets. Note, $x$-axis is in logarithmic scale.} 
	\label{fig:acuracy_vs_pgdsteps}    
	\vspace{-0.3cm}
\end{figure*}

We train models on MNIST, Fashion-MNIST and CIFAR-10 datasets respectively, using NT, FAT, PAT, TRADES and \sads (Algorithm~\ref{alg:ProposedAlgorithm}) training methods. Models are trained for 50, 50 and 100 epochs on MNIST, Fashion-MNIST and CIFAR-10 datasets respectively. For \sads, we set the hyper-parameter \DP and \DR to (0.8, 0.5), (0.8, 0.75) and (0.5, 0.5) for MNIST, Fashion-MNIST and CIFAR-10 datasets respectively. Table~\ref{table:whitebox_mnist}, ~\ref{table:whitebox_fmnist} and~\ref{table:whitebox_cifar} shows the performance of these models against single-step and multi-step attacks in white-box setting, rows represent the training method and columns represent the attack generation method. It can be observed that models trained using FAT are not robust against multi-step attacks. Whereas, models trained using PAT, TRADES and \sads are robust against both single-step and multi-step attacks. Unlike PAT and TRADES, the proposed \sads method is a single-step adversarial training method.\\ \\
\noindent\textbf{PGD attack with large steps}: Engstrom~\etal\cite{engstrom2018evaluating} demonstrated that the performance of models trained using certain adversarial training methods degrade significantly with increase in the number of steps of PGD attack. In order to verify that such behavior is not observed in models trained using \sads, we obtain the plot of classification accuracy on PGD test-set versus steps of PGD attack. Fig.~\ref{fig:acuracy_vs_pgdsteps} shows these plots obtained for models trained using PAT and \sads on MNIST, Fashion-MNIST and CIFAR-10 datasets respectively. It can be observed that the accuracy of models on PGD test set initially decreases slightly and then saturates. Even for PGD attack with large steps, there is no significant degradation in the performance of models trained using PAT and \sads methods. In supplementary document, we show the effect of hyper-parameters of the proposed training method.
\begin{figure*}[t!]
\centering    
	\includegraphics[width=0.90\linewidth]{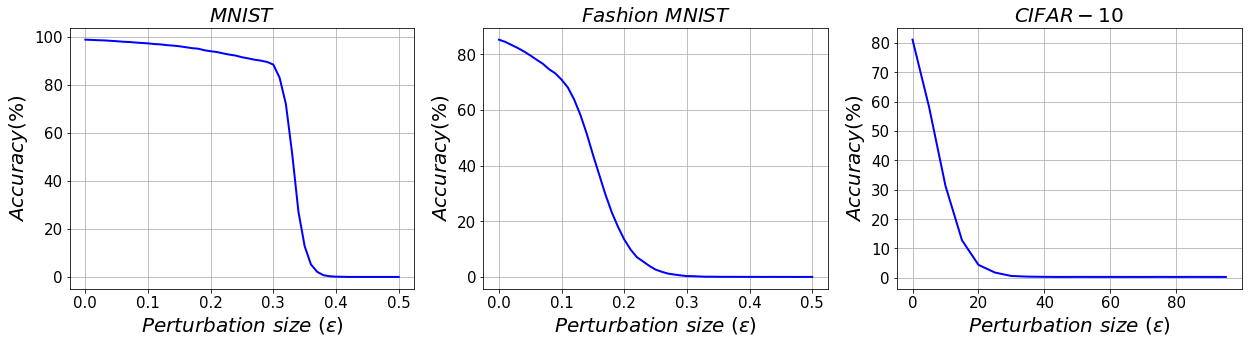}
    \vspace{-0.3cm}
	\caption{Plot of accuracy versus perturbation size of PGD attack, obtained for models trained using \sads. It can be observed that the accuracy of the model is zero for PGD attack with large perturbation size.} 
	\label{fig:acc_vs_pgdeps}   
	\vspace{-0.3cm}
\end{figure*}
\begin{table}[h!]
\caption{Black-box setting: Performance of models trained on MNIST, Fashion-MNIST and CIFAR-10 datasets using different training method, against adversarial attacks in black-box setting. Source models are used for generating adversarial samples, and the target models are tested on these generated adversarial samples.}
\label{table:blackboc_attack}
\centering
 \setlength\tabcolsep{3pt}
\resizebox{1.0\linewidth}{!}{
\begin{tabular}{|l|l|l|l|l|l|}
\hline
\multicolumn{6}{|c|}{\textbf{MNIST}}                                                             \\ \hline
\multicolumn{2}{|c|}{\multirow{2}{*}{Source Model}} & \multicolumn{4}{c|}{Target Model} \\ \cline{3-6} 
\multicolumn{2}{|c|}{}                              & NT   & FAT  & PAT  & \sads  \\ \hline \hline
\multirow{2}{*}{Model-A}             & FGSM ($\epsilon$=0.3)             & 29.09	& 79.49 & 96.01 & 95.06    \\ \cline{2-6} 
                              & MI-FGSM ($\epsilon$=0.3, steps=40)       & 10.69	& 72.44	& 95.83 & 94.80    \\ \hline
\multirow{2}{*}{Model-B}             & FGSM ($\epsilon$=0.3)             & 28.13	& 72.39	& 96.15 & 95.11   \\ \cline{2-6}
                              & MI-FGSM ($\epsilon$=0.3, steps=40)       & 12.32	& 70.79	& 95.97 & 94.81    \\ \hline
\multicolumn{6}{|c|}{\textbf{Fashion-MNIST}}                                                     \\ \hline
\multirow{2}{*}{Model-A}             & FGSM ($\epsilon$=0.1)             & 36.66    & 88.26 & 81.32	& 80.86    \\ \cline{2-6} 
                              & MI-FGSM ($\epsilon$=0.1, steps=40)       & 33.04	& 88.36	& 81.20	& 80.68    \\ \hline
\multirow{2}{*}{Model-B}             & FGSM ($\epsilon$=0.1)             & 39.03	& 85.40	& 80.01 & 78.94    \\ \cline{2-6} 
                              & MI-FGSM ($\epsilon$=0.1, steps=40)       & 38.01	& 84.72	& 79.84	& 78.59   \\ \hline
\multicolumn{6}{|c|}{\textbf{CIFAR-10}}                                                     \\ \hline
\multirow{2}{*}{VGG-11}             & FGSM ($\epsilon$=8/255)           & 48.46 & 78.70 & 78.12 & 77.97 \\ \cline{2-6} 
                              & MI-FGSM ($\epsilon$=8/255, steps=7)     & 31.61 & 76.35 & 78.36 & 77.95 \\      \hline
DenseNet-             & FGSM ($\epsilon$=8/255)                         & 39.58 & 86.90 & 80.29 & 80.06 \\ \cline{2-6} 
BC-100                              & MI-FGSM ($\epsilon$=8/255, steps=7)     & 28.50 & 86.42 & 80.42 & 80.28      \\ \hline                     
\end{tabular}}
\end{table}
\subsection{Performance in Black-box setting}
\label{subsec:exp_black_box_setting}
In this subsection, we show the performance of models trained using different training methods against adversarial attacks in black-box setting. Typically, a substitute model (source model) is trained on the same task using normal training method, and this trained substitute model is used for generating adversarial samples. The generated adversarial samples are transferred to the deployed model (target model). We use FGSM and MI-FGSM methods for generating adversarial samples, since samples generated using these methods show good transfer rates~\cite{dong2018boosting}. Table~\ref{table:blackboc_attack} shows the performance of models trained using different methods, in black-box setting. It can be observed that the performance of models trained using  PAT and \sads in black-box setting is better than that in white-box setting. Further, it can be observed that the performance of  models trained on MNIST and CIFAR-10 datasets using FAT is worse in black-box setting than compared in white-box setting. Please refer to supplementary file for details on network architecture of source models.
\begin{table*}[]
\caption{\textbf{DeepFool and C\&W attacks:} Performance of models trained  using different training methods against DeepFool and C\&W attacks. These attack methods measure the robustness of the model based on the average $l_2$ norm of the generated perturbations, higher the better. Success defines the percentage of samples of test set that has been misclassified. Note that, for models trained using PAT and \sads, perturbations with relatively large $l_2$ norm is required to fool the classifier.}
\vspace{-0.1cm}
\centering
\label{table:deepfool_cw}
\setlength\tabcolsep{3pt}
\resizebox{1.0\linewidth}{!}{
\begin{tabular}{|l|r|r|r|r|r|r|r|r|r|r|r|r|}
\hline
\multirow{3}{*}{Method} & \multicolumn{4}{c|}{MNIST}                              & \multicolumn{4}{c|}{F-MNIST}                            & \multicolumn{4}{c|}{CIFAR-10}                           \\ \cline{2-13} 
                              & \multicolumn{2}{c|}{DeepFool} & \multicolumn{2}{c|}{CW} & \multicolumn{2}{c|}{DeepFool} & \multicolumn{2}{c|}{CW} & \multicolumn{2}{c|}{DeepFool} & \multicolumn{2}{c|}{CW} \\ \cline{2-13} 
                              & Success       & Mean $l_2$       & Success    & Mean $l_2$    & Success       & Mean $l_2$       & Success    & Mean $l_2$    & Success       & Mean $l_2$       & Success    & Mean $l_2$    \\ \hline
NT  & 99.35 & 1.837 & 100 & 1.659 &93.73 & 0.796  & 100 & 0.709 & 96 & 0.20 & 100 & 0.12 \\ \hline
FAT & 99.37 & 1.455 & 100 & 0.798 &93.11 & 1.514  & 100 & 1.167 & 96 & 0.25 & 100 & 0.10 \\ \hline
PAT & 85.68 & 4.633 & 99  & 2.779 &90.29 & 2.635  & 100 & 1.572 & 92 & 1.22 & 100 & 0.88 \\ \hline
SADS& 95.89 & 3.692 & 100 & 2.321 &90.68 & 2.305  & 100 & 1.308 & 93 & 0.97 & 100 & 0.71 \\ 
    &$\pm$0.06  &$\pm$0.033 & 0$\pm$  & $\pm$0.027 &$\pm$0.26  & $\pm$0.102  & $\pm$0   & $\pm$0.188 & $\pm$0.32  & $\pm$0.043 & $\pm$0   & $\pm$0.014 \\ \hline
\end{tabular}}
\end{table*}

\begin{table}[h!]
\vspace{-0.3cm}
\centering
\caption{Comparison of training time per epoch of models trained on MNIST and CIFAR-10 datasets respectively, obtained for different training methods.}
\vspace{-0.1cm}
\label{table:training_time}
\begin{tabular}{lcc}\hline 
\multirow{ 2}{*}{\textbf{Method}} & \multicolumn{2}{c}{Training time per epoch (sec.)}\\ \cline{2-3}
                & MNIST & CIFAR-10                                  \\ \hline \hline
    NT          & $\sim2.7$   & $\sim104$                \\
    FAT         & $\sim4.1$   & $\sim159$               \\
    PAT         & $\sim53$  & $\sim820$                \\
    TRADES      & $\sim104$   & $\sim1558$                \\
    \sads       & $\sim4.3$   & $\sim187$                \\ \hline
\end{tabular}
\vspace{-0.3cm}
\end{table}
\subsection{Performance against DeepFool and C\&W attacks}
\label{subsec:exp_deepfool_cw_attack}
DeepFool~\cite{deepfool-cvpr-2016} and C\&W~\cite{robustness-arxiv-2016} attacks generate adversarial perturbations with minimum $l_2$ norm, that is required to fool the classifier. These methods measure the robustness of the model in terms of the average $l_2$ norm of the generated adversarial perturbations for the test set. For an undefended model, adversarial perturbation with small $l_2$ norm is enough to fool the classifier. Whereas for robust models, adversarial perturbation with relatively large $l_2$ norm is required to fool the classifier. Table~\ref{table:deepfool_cw}, shows the performance of models trained using NT, FAT, PAT and \sads methods, against DeepFool and C\&W attacks. It can be observed that models trained using PAT and \sads  have relatively large average $l_2$ norm. Whereas, for models trained using NT and FAT have small average $l_2$ norm. 
\subsection{Sanity tests}
\label{subsec:exp_obfuscated_gradeint_tests}
\vspace{-0.1cm}
We perform sanity tests described in~\cite{carlini2019evaluating} to verify whether models trained using \sads are adversarially robust and are not exhibiting obfuscated gradients. We perform following sanity tests:
\begin{itemize}[noitemsep]
\item \textit{Iterative attacks should perform better than non-iterative attacks}  
\item \textit{White-box attacks should perform better than black-box attacks}
\item \textit{Unbounded attacks should reach 100\% success}
\item \textit{Increasing distortion bound should increase attack success rate}
\end{itemize}
Models trained using \sads pass above tests. From table~\ref{table:whitebox_mnist},~\ref{table:whitebox_fmnist} and ~\ref{table:whitebox_cifar}, it can be observed that iterative attacks (IFGSM and PGD) are stronger than non-iterative attack (FGSM) for models trained using  \sads. Comparing results in Tables~\ref{table:whitebox_mnist},~\ref{table:whitebox_fmnist} and ~\ref{table:whitebox_cifar} with results in Table~\ref{table:blackboc_attack}, it can be observed that white-box attacks are stronger than black-box attacks for models trained using \sads. Fig.~\ref{fig:acc_vs_pgdeps} shows the accuracy plot for the model on test set versus perturbation size of PGD attack, obtained for models trained using \sads. It can be observed that the model's accuracy falls to zero for large perturbation size ($\epsilon$). From Fig.~\ref{fig:acc_vs_pgdeps}, it can be observed that  PGD attack success rate (attack success rate is equal to (100 - model's accuracy)\%) increases with increase in the distortion bound (perturbation size) of the attack. 
\subsection{Time Complexity}
\label{subsec:Time Complexity}
\vspace{-0.1cm}
In order to quantify the complexity of different training methods, we measure training time per epoch (seconds) for models trained using different training methods. Table~\ref{table:training_time} shows the training time per epoch for models trained on MNIST and CIFAR-10 datasets respectively. Note that the training time of \sads and FAT is of the same order. The increase in the training time for PAT  and TRADES is due to their iterative nature of generating adversarial samples. We ran this timing experiment on a machine with NVIDIA Titan Xp  GPU, with no other jobs on this GPU.
\section{Conclusion}
\label{sec:Conclusion}
\vspace{-0.1cm}
In this work, we have demonstrated that models trained using single-step adversarial training methods learn to prevent the generation of adversaries due to over-fitting of the model during the initial stages of training. To mitigate this effect, we have proposed a novel single-step adversarial training method with dropout scheduling. Unlike existing single-step adversarial training methods, models trained using the proposed method achieves robustness not only against single-step attacks but also against multi-step attacks. Further, the performance of models trained using the proposed method is on par with models trained using multi-step adversarial training methods, and is much faster than multi-step adversarial training methods.\\
\\\textbf{Acknowledgment}: This work was supported by  Uchhatar Avishkar Yojana (UAY) project (IISC\_010), MHRD, Govt. of India.
{\small
\bibliographystyle{ieee_fullname}
\bibliography{main}
}
\clearpage
\renewcommand{\thesection}{S~\arabic{section}}
\renewcommand{\thesubsection}{\thesection.\arabic{subsection}}
\setcounter{section}{0}
\title{Supplementary}
\author{}
\date{}
\maketitle
\section{Contents}
\begin{itemize}
    \item Section~\ref{sec:network_architecture}: Network architecture.
    \item Section~\ref{sec:adversarial-sample-generation-methods}: Adversarial training and Attack generation methods.
    \item Section~\ref{sec:additional_plots}: Additional plots to illustrate over-fitting effect during single-step adversarial training.
    \item Section~\ref{sec:exp_effect_of_hyperparameter}: Effect of hyper-parameters
    \item Section~\ref{sec:comparison_with_eat}: Comparison with EAT
    \item Section~\ref{sec:trend_train_val}: Trend of $R_{\epsilon}$, training and validation loss during \sads.
\end{itemize}
\section{Network Architecture}
\label{sec:network_architecture}
Network architecture of  models used for MNIST and Fashion-MNIST datasets are shown in Table~\ref{table:architecure}. Model-A and Model-B are used for generating adversarial samples in black-box setting.
\begin{table*}[bh!]
    \caption{Architecture of networks used for MNIST and Fashion-MNIST datasets.}
    \vspace{-0.3cm}
    \centering
    \label{table:architecure}
    \begin{tabular}{c|c|c|c|cc} \hline\hline
    \textbf{LeNet+} & \textbf{Model-A} & \textbf{Model-B} & \textbf{Model-C} & \textbf{Model-D} \\ \hline
    Conv(32,5,5) + Relu & Conv(64,5,5) + Relu  & Dropout(0.2)         & Conv(128,3,3) + Tanh   & \multirow{2}{*}{$\Big\{$} FC(300) +Relu \multirow{2}{*}{$\Big\}\times4$}\\
    MaxPool(2,2) & Conv(64,5,5) + Relu  & Conv(64,8,8)  + Relu & MaxPool(2,2)           & Dropout(0.5)  \\
    Conv(64,5,5) + Relu & Dropout(0.25)        & Conv(128,6,6) + Relu & Conv(64,3,3) + Tanh    & FC + Softmax  \\
    MaxPool(2,2) & FC(128) + Relu       & Conv(128,5,5) + Relu & MaxPool(2,2)           &   \\ 
    FC(1024) + Relu & Dropout(0.5)        & Dropout(0.5)        & FC(128) + Relu         &   \\
    FC + Softmax    & FC + Softmax         & FC + Softmax         & FC + Softmax           &   \\ \hline\hline
    \end{tabular}
\end{table*}
\begin{figure*}[h!]
    \centering
	\includegraphics[width=0.99\linewidth]{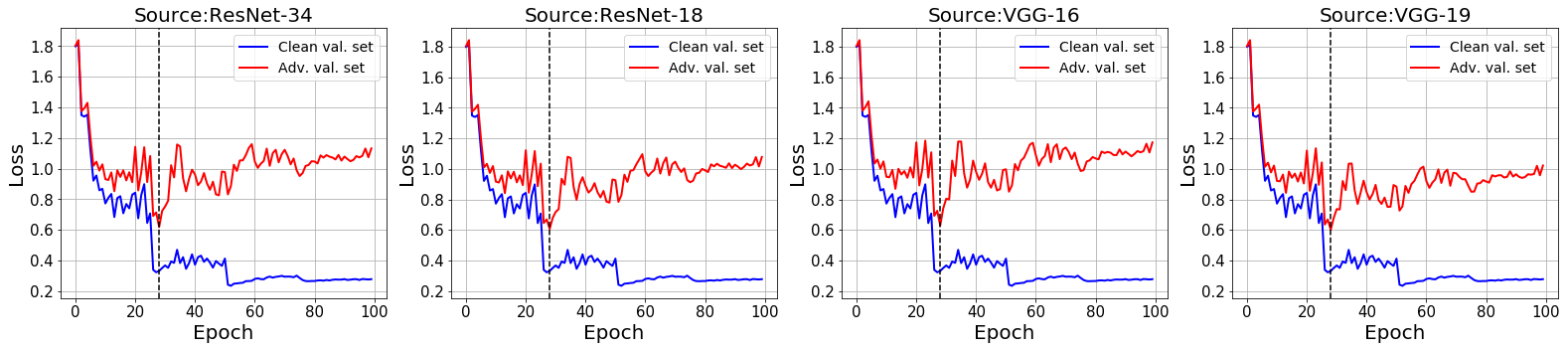}
	\vspace{-0.4cm}
	\caption{\textbf{Single-step adversarial training}: Trend of validation loss during single-step adversarial training, obtained for ResNet-34 trained on CIFAR-10 dataset. Adversarial validation set is generated using column-1: ResNet-34, column-2: ResNet-18, column-3: VGG-16 and column-4: VGG-19.} 
	\vspace{-0.4cm}
	\label{fig:trend_fat_cifar10_additional}  
\end{figure*}
\begin{figure*}[h!]
    \centering
	\includegraphics[width=0.99\linewidth]{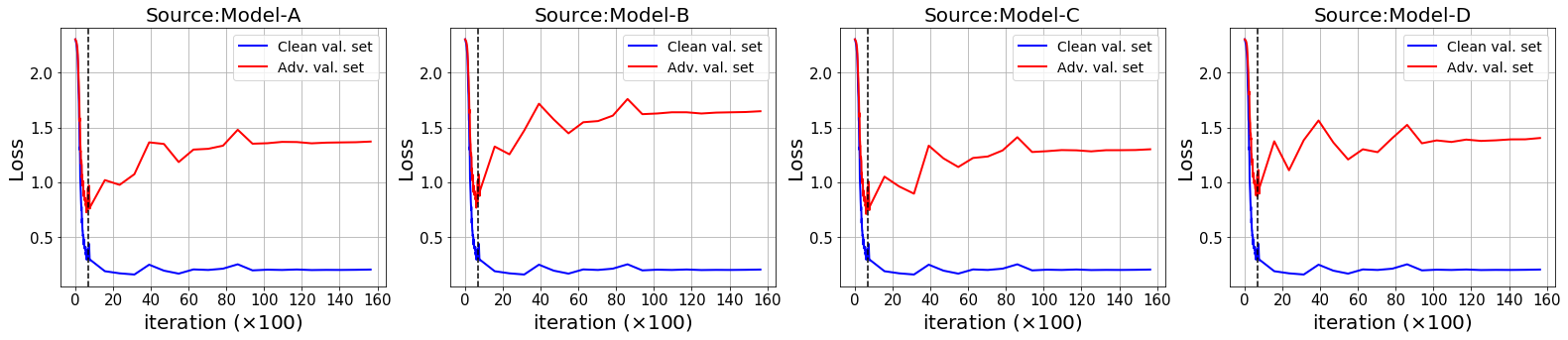}
	\vspace{-0.4cm}
	\caption{\textbf{Single-step adversarial training}: Trend of validation loss during single-step adversarial training, obtained for LeNet+ trained on MNIST dataset. Adversarial validation set is generated using column-1: Model-A, column-2: Model-B, column-3: Model-C and column-4: Model-D.} 
	\vspace{-0.4cm}
	\label{fig:trend_fat_mnist_additional}  
\end{figure*}
\section{Adversarial training and Attack generation methods}
\label{sec:adversarial-sample-generation-methods}
\subsection{Adversarial Sample Generation Methods}
In this subsection, we discuss the formulation of adversarial attacks.\\
\textbf{Fast Gradient Sign Method (FGSM)}: Non-iterative attack method proposed by~\cite{explainingharnessing-iclr-2015}. This method generates $l_{\infty}$ norm bounded adversarial perturbation based on the linear approximation of loss function.
\begin{equation}
x^* = x + \epsilon. sign(\nabla_{x}J(f(x;\theta),y_{true}))
\end{equation}
\textbf{Iterative Fast Gradient Sign Method (IFGSM)}~\cite{physicalworld-arxiv-2016}: Iterative version of FGSM attack. At each iteration, adversarial perturbation of small step size ($\alpha$) is added to the image. In our experiments, we set $\alpha=\epsilon/steps$.
\begin{eqnarray}
x^{0} & = & x\\
x^{N+1} & = & x^{N} + \alpha . sign\big(\nabla_{x^{N}}J(f(x^{N};\theta), y_{true})\big)
\end{eqnarray}
\textbf{Projected Gradient Descent (PGD)}~\cite{madry-iclr-2018}: Initially, a small random noise sampled from Uniform distribution ($U$) is added to the image. Then at each iteration,  perturbation of small step size ($\epsilon_{step}$) is added to the image, and followed by re-projection.
\begin{eqnarray}
x^{0} & = & x + U\big(-\epsilon_{step},\epsilon_{step},shape(x)\big) \\
x^{N+1} & = & x^{N} + \epsilon_{step} . sign\big(\nabla_{x^{N}}J(f(x^{N};\theta), y_{true})\big) \\
x^{N+1} & = & clip\big(x^{N+1},min=x-\epsilon,max=x+\epsilon\big) 
\end{eqnarray}
\\
\\\textbf{Momentum Iterative Fast Gradient Sign Method (MI-FGSM)~\cite{dong2018boosting}}: Introduces a momentum term into the IFGSM formulation. Here, $\mu$ represents the momentum term. $\alpha$ represents step size and is set to $\epsilon/steps$. 
\begin{eqnarray}
x^{0} & = & x\\
g^{N+1} &=& \mu.g^{N} +  \frac{\nabla_{x^{N}}J(f(x^{N};\theta), y_{true})}{||\nabla_{x^{N}}J(f(x^{N};\theta), y_{true})||_1}\\
x^{N+1} & = & x^{N} + \alpha . sign\big(g^{N+1}\big)
\end{eqnarray}
\subsection{Adversarial Training Methods}
\label{sec:appendex_training_methods}
In this subsection we explain the existing adversarial training methods.
\\\textbf{FGSM Adversarial Training (FAT)}: During training, at each iteration a portion of clean samples in the mini-batch are replaced with their corresponding adversarial samples generated using the model being trained. Fast Gradient Sign Method (FGSM) is used for generating these adversarial samples.
\\\textbf{Ensemble Adversarial Training (EAT)}~\cite{ensembleAT-iclr-2018}: At each iteration a portion of clean samples in the mini-batch are replaced with their corresponding adversarial samples. These adversarial samples are generated by the model being trained or by one of the model from the fixed set of pre-trained models. Table~\ref{table:ensemble_setup} shows the setup used for EAT method.
\\\textbf{PGD Adversarial Training (PAT)}: Multi-step adversarial training method proposed by~\cite{madry-iclr-2018}. At each iteration all the clean samples in the mini-batch are replaced with their corresponding adversarial samples generated using the model being trained. Projected Gradient Descent (PGD) method is used for generating these samples.
\\\textbf{TRADES}: Multi-step adversarial training method proposed by~\cite{trades_icml}. The method proposes a regularizer that encourages the output of the network to be smooth. The training mini-batches contain clean and their corresponding adversarial samples. These adversarial samples are generated using Projected Gradient Descent with modified loss function.
\section{Additional plots to illustrate over-fitting effect}
\label{sec:additional_plots}
In the main paper, we showed over-fitting effect during training of LeNet+ on MNIST dataset using single-step adversarial training. Fig.~\ref{fig:trend_fat_cifar10_additional} shows the plot of validation loss, obtained for ResNet-34 trained on CIFAR-10 dataset using single-step adversarial training. We observe over-fitting effect even when model with different architecture is used for generating adversarial validation set. Fig.~\ref{fig:trend_fat_mnist_additional} shows the validation loss obtained for LeNet+ trained on MNIST dataset using single-step adversarial training. Normally trained models with different architecture are used for generating adversarial validation set.
\begin{figure}[h!]
        \centering    
        \includegraphics[width=0.99\linewidth]{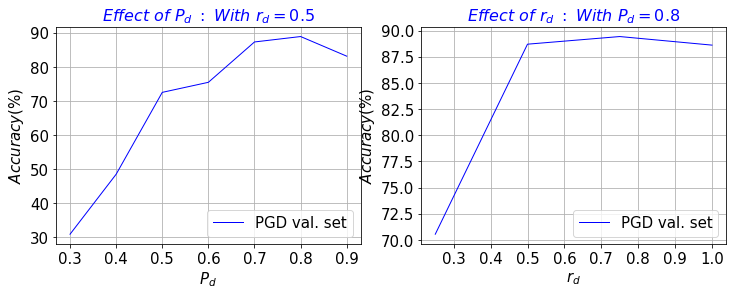}
        \caption{Effect of hyper-parameter \DP and \DR of SADS} 
        \label{fig:effect_of_hyperparameters_DP_DR} 
\end{figure}
\begin{figure*}[t!]
    \centering
	\includegraphics[width=0.99\linewidth]{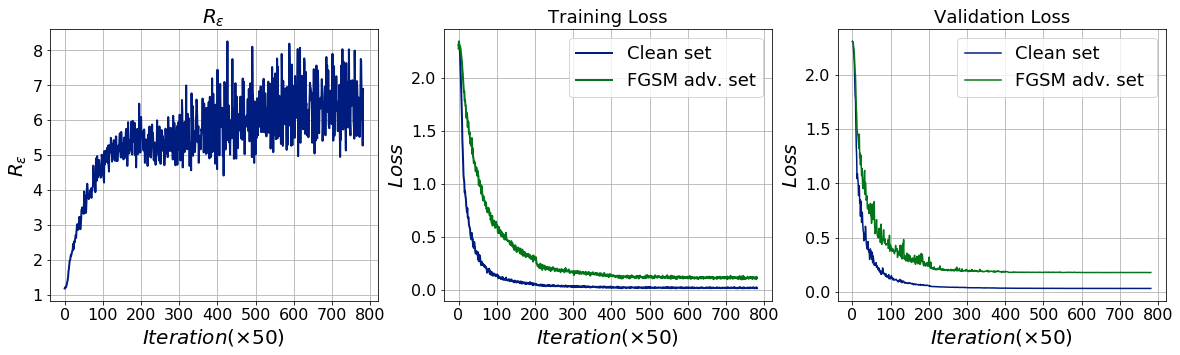}
	\vspace{-0.4cm}
	\caption{\textbf{MNIST}: Trend of $R_{\epsilon}$, training loss, and validation loss during \sads training method, obtained for LeNet+ trained on MNIST dataset. Column-1: plot of $R_{\epsilon}$ versus iteration. Column-2: training loss versus iteration. Column-3: validation loss versus iteration. Note that, for the entire training duration $R_{\epsilon}$ does not decay, and no over-fitting effect can be observed.} 

	\label{fig:trends_during_sads_mnist}  
	 \centering
	\includegraphics[width=0.99\linewidth]{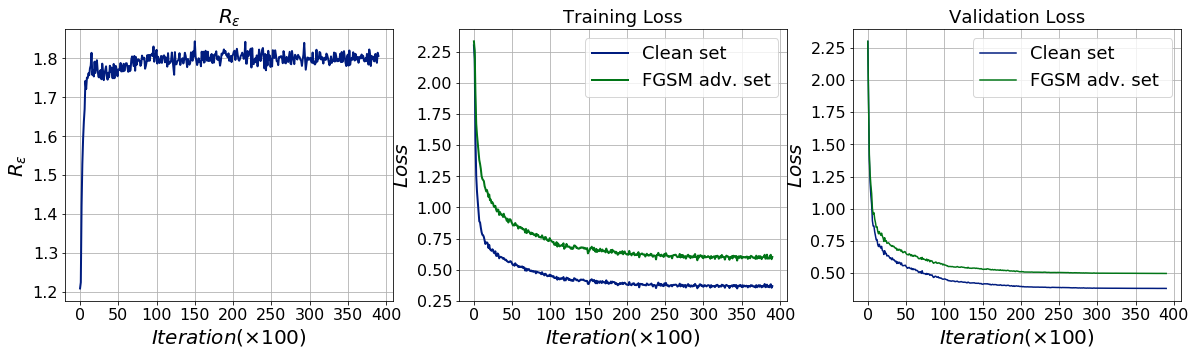}
	\caption{\textbf{Fashion-MNIST}: Trend of $R_{\epsilon}$, training loss, and validation loss during \sads training method, obtained for LeNet+ trained on Fashion-MNIST dataset. Column-1: plot of $R_{\epsilon}$ versus iteration. Column-2: training loss versus iteration. Column-3: validation loss versus iteration. Note that, for the entire training duration $R_{\epsilon}$ does not decay, and no over-fitting effect can be observed.} 
	\label{fig:trends_during_sads_fmnist}  
\end{figure*}
\section{Effect of Hyper-Parameters}
\label{sec:exp_effect_of_hyperparameter}
In order to show the effect of hyper-parameters, we train LeNet+ shown in table~\ref{table:architecure} on MNIST dataset, using \sads with different hyper-parameter settings. Validation set accuracy of the model for PGD attack ($\epsilon=0.3$ and $steps=40$) is obtained for each hyper-parameter setting with one of them being fixed and the other being varied.\\
 \textbf{Effect of hyper-parameter \DP}: The hyper-parameter \DP defines the initial dropout probability applied to all dropout layers. We train LeNet+ on MNIST dataset, using the proposed method for different initial dropout probability \DP. Column-1 of Fig.~\ref{fig:effect_of_hyperparameters_DP_DR} shows the effect of varying dropout probability from $0.3$ to $0.9$. It can be observed that the robustness of the model to multi-step attack initially increases with the increase in the value of \DP (\DP$< 0.8$), and further increase in \DP causes the model's robustness to decrease, and this is due to under-fitting.
\\\textbf{Effect of hyper-parameter \DR}: The hyper-parameter \DR decides the iteration at which dropout probability reaches zero and is expressed in terms of maximum training iteration. Column-2 of Fig.~\ref{fig:effect_of_hyperparameters_DP_DR} shows the effect varying \DR from $1/4$ to $1$. It can be observed that for \DR $<0.5$, there is degradation in the robustness of the model against multi-step attacks. This is because, during the initial stages of training, learning rate is high and the model can easily over-fit to adversaries generated by single-step method.
\section{Comparison with Ensemble Adversarial Training}
\label{sec:comparison_with_eat}
We train WideResNet-28-10~\cite{BMVC2016_87} on CIFAR-10~\cite{cifar_10_dataset} dataset using EAT and \sads. Table~\ref{table:ensemble_setup} shows the setup used for EAT. Pre-trained models are used for generating adversarial samples during EAT. Table~\ref{table:whitebox_cifar_eat_comparison} shows the recognition accuracy of models trained using EAT and \sads in white-box attack setting. It can be observed that the model trained using \sads is robust to both single-step (FGSM) and multi-step attacks (PGD), whereas models trained using EAT are susceptible to multi-step attack.
\section{\sads: Trend of $R_{\epsilon}$, training and validation loss}
\label{sec:trend_train_val}
Fig.~\ref{fig:trends_during_sads_mnist} and ~\ref{fig:trends_during_sads_fmnist}  show the trend of $R_{\epsilon}$, training and validation loss, obtained for models trained using \sads. It can be observed that for the entire training duration $R_{\epsilon}$ does not decay and no over-fitting effect can be observed.
\begin{table}[h]
\caption{Setup used for Ensemble Adversarial Training.}
    \centering
    \label{table:ensemble_setup}
    \setlength\tabcolsep{3pt}
    \resizebox{1.0\linewidth}{!}{
    \begin{tabular}{ccc}\hline\hline
                                             & \textbf{Network to be trained}      & \textbf{Pre-trained Models }    \\ \hline\hline
                                 & WRN-28-10 (Ens-A)               & WRN-28-10, ResNet-34            \\ 
CIFAR-10                         & WRN-28-10 (Ens-B)               & WRN-28-10, VGG-19               \\ 
                                 & WRN-28-10 (Ens-C)               & ResNet-34, VGG-19              \\ \hline\hline
    \end{tabular}
    }
\end{table}

\begin{table}[h!]
    \centering
    \caption{\textbf{CIFAR-10: White-Box attack.} Classification accuracy (\%) of models trained on CIFAR-10 dataset using different training methods.  For all attacks $\epsilon$=8/255 is used and for PGD attack $\epsilon_{step}$=2/255 and $steps$=7 is used.}
    \label{table:whitebox_cifar_eat_comparison}
    \begin{tabular}{lcrrr} \hline \hline
    \multicolumn{2}{l}{\textbf{Training}} & \multicolumn{3}{c}{\textbf{Attack Method}} \\ \cline{4-5}
    \multicolumn{2}{l}{\textbf{Method}} & \textbf{Clean}  & \textbf{FGSM}   & \textbf{PGD-7}         \\
     \hline \hline
    \multicolumn{2}{l}{EAT Ens-A}   & 92.92      & 59.56      & 19.21     \\
    \multicolumn{2}{l}{EAT Ens-B}   & 92.75      & 63.40      & 5.34     \\
    \multicolumn{2}{l}{EAT Ens-C}   & 93.11      & 59.74      & 12.03     \\\hline
    \multicolumn{2}{l}{\sads}       & 82.01      & 51.99      & 45.66     \\ 
    \multicolumn{2}{l}{}            & $\pm$0.06  & $\pm$1.02  & $\pm$1.26     \\ \hline \hline
    \end{tabular}
\end{table}

\end{document}